\documentclass[lettersize,journal]{IEEEtran}
\usepackage{amsmath,amsfonts}
\usepackage{algorithmic}
\usepackage{algorithm}
\usepackage{array}
\usepackage[caption=false,font=normalsize,labelfont=sf,textfont=sf]{subfig}
\usepackage{textcomp}
\usepackage{stfloats}
\usepackage{url}
\usepackage{verbatim}
\usepackage{graphicx}
\usepackage{cite}
\usepackage{xcolor}
\usepackage{bm}
\usepackage{adjustbox}
\usepackage{tabularx}
\usepackage{multirow}
\usepackage{tikz}
\usepackage{algorithm}
\usepackage{enumitem}
\usepackage{booktabs}
\usepackage{graphicx}
\usepackage{amsmath}
\usepackage{array}
\usepackage{colortbl}
\usepackage{setspace}

\usepackage[labelformat=simple]{subcaption}  
\usepackage{titletoc}
\usepackage[font=small]{caption}

\usepackage{soul}

\begin{document}

\title{LiMoDE: Rethinking Lifelong Robot Manipulation from a Mixture-of-Dynamic-Experts Perspective}

\author{Zhihao Gu\textsuperscript{*}, Lin Wang$^\dagger$\\
School of EEE, Nanyang Technological University
    \thanks{\textsuperscript{*}Work done when Zhihao Gu was a Postdoctoral Research Fellow at NTU.}
    \thanks{$^\dagger$Corresponding author.}
}



\maketitle

\begin{abstract}
Building a generalist robot that can leverage prior knowledge for continuous task adaptation remains a significant challenge. 
Previous works alleviate the catastrophic forgetting problem by parameter-efficient fine-tuning for single-task adaptation. However, they fail to extract reusable skills and model the interaction with other skills effectively. Recent works try to address these issues by learning prompts. Differently, this paper presents an architectural perspective on the Lifelong Mixture of Dynamic Experts (\textit{LiMoDE}), a novel two-stage learning scheme for lifelong robot manipulation. 
Specifically, a dynamic MoE structure is first proposed in the multi-task pre-training stage to learn prior knowledge, where a varied number of heterogeneous experts are activated based on the motion information to address different short-term manipulations. 
Subsequently, in the task adaptation stage, we design a lifelong MoE adaptation mechanism 
that learns lifelong experts and dynamically combines them with frozen ones for new tasks, facilitating the knowledge transfer during adaptation. 
The proposed \textit{LiMoDE} is evaluated on both the simulated lifelong learning benchmark and real-world tasks. 
Extensive experiments demonstrate its effectiveness in achieving superior performance and strong lifelong adaptation by introducing a moderate number of additional trainable parameters and inference overhead.
\end{abstract}
\vspace{-4pt}


\section{Introduction}\label{intro}
\IEEEPARstart{R}{ecently}, the robotics field has made vital progress in learning generalist policies to solve complex tasks~\cite{brohan2022rt,zitkovich2023rt,black2410pi0,gu2026learning,jang2022bc}. Despite the success, they struggle with catastrophic forgetting issues during continual task adaptation~\cite{yao2025think,roy2025m2distill,lee2025policy}, where performance on previously seen tasks can not be well retained.
 
Researchers have developed robots to continuously learn and adapt throughout their lifetime, known as lifelong learning~\cite{de2021continual}, and various approaches~\cite{gao2021cril,zhu2022bottom,liu2023libero,wan2024lotus} show promising results in imitation learning from sequentially arrived tasks while alleviating forgetting on previously learned tasks. 
These approaches can be further categorized into three main types: replay, regularization, and architectural methods. Experience replay~\cite{rolnick2019experience,gao2024efficient} explicitly utilizes previous data, but faces large storage space and computational overhead for retaining old samples. Regularization-based methods~\cite{li2017learning,zenke2017continual} restrict the update of parameters to maintain existing knowledge when learning new tasks. 
In contrast, architectural methods~\cite{liu2023tail,schmied2023learning,yao2025think} integrate learnable modules in single-task adaptation to address catastrophic forgetting via parameter-efficient fine-tuning (PEFT)~\cite{he2021towards}. More recently, TAIL~\cite{liu2023tail} explores learning task-specific parameters by low-rank adaptation (LoRA)~\cite{hu2022lora}, and retrieves corresponding adapters. 

\begin{figure}[t]
  \centering
  \includegraphics[width=0.9\linewidth]{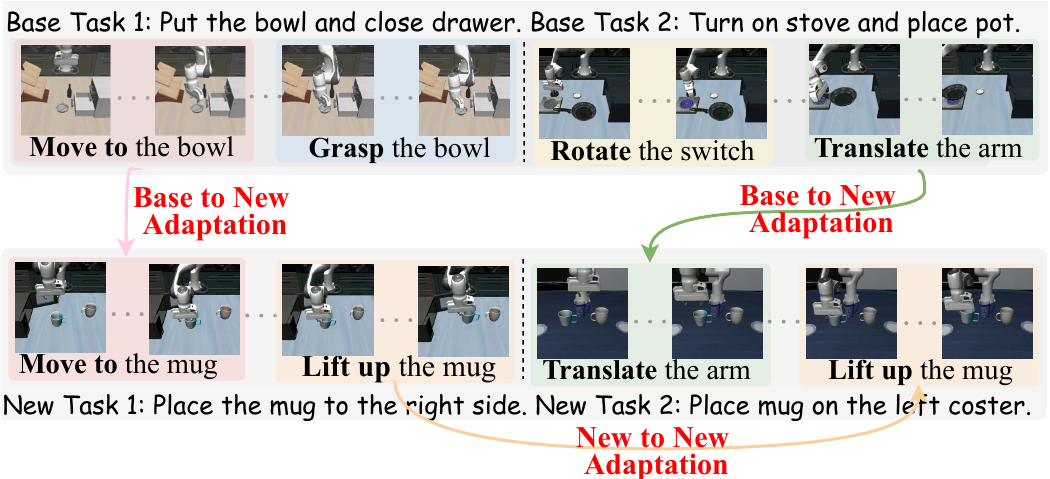}
  \vspace{-5pt}
   \caption{\textbf{Motivation of our framework.} Base tasks contain various short-term actions, and skills in them (highlighted in bold) can be adapted to complete new tasks. New tasks contain shared knowledge, and interaction between them is beneficial for continual adaptation.}
   \label{fig2}
   \vspace{-8pt}
\end{figure}

Although these approaches achieve good results, they fail to exploit the underlying interactions among shared knowledge across tasks, resulting in suboptimal knowledge reuse and slow adaptation.
To bridge this gap, we draw inspiration from two observations illustrated in Fig.~\ref{fig2}. 
First, base tasks contain various primitive short-term actions, and skills within them can be adapted to conduct similar actions in new tasks. For instance, while 
actions like ``\textit{Move to the bowl}'' in the base task and ``\textit{Move to the mug}'' in the new task are semantically distinct, they share a common skill ``\textit{Move to}''.
\IEEEpubidadjcol
Second, new tasks inherently contain overlapping information, and modeling their correlations can facilitate cross-task knowledge transfer.
Therefore, \textit{our objective is to extract those reusable skills and then model the interactions in a unified framework}.

To this end, this paper introduces the Lifelong Mixture of Dynamic Experts (\textit{\textbf{LiMoDE}}), a novel two-stage learning scheme, for lifelong robot manipulation. \textit{In the multi-task pre-training stage}, a dynamic MoE structure is constructed to learn a set of reusable skills from base tasks. Specifically, these skills are represented by the combination of a base expert and some heterogeneous experts, the number of which is dynamically adjusted conditioned on the motion intensity, increasing with large motion and decreasing conversely. To encourage diversity and sparsity, we adopt a lightweight router-decorrelation regularizer. \textit{In the lifelong learning stage}, we devise a lifelong MoE adaptation mechanism that progressively learns low-rank experts and dynamically assembles them via a lightweight router network, facilitating cross-task interaction and knowledge transfer. A replay strategy is then applied to the router for expert retrieval, mitigating forgetting on previously learned tasks. Evaluations in simulation and the real world demonstrate consistent improvements over previous methods in continual adaptation. To summarize, our work contributes:
\begin{itemize}
    \item \textbf{The Lifelong Mixture of Dynamic Experts} (\textit{\textbf{LiMoDE}}) for lifelong robot learning. It enables reusable skill learning and cross-task interaction in a unified framework.
    \item \textbf{A Dynamic MoE structure} for modelling reusable skills as an assembly of a varied number of experts. A lightweight router-decorrelation regularizer is adopted to encourage diverse and sparse routing.
    \item \textbf{A lifelong MoE adaptation mechanism} for cross-task interaction. It progressively learns experts and assembles them, mitigating forgetting by a replay strategy.
    \item \textbf{State-of-the-art performance} for LIBERO benchmark, enhancing adaptation by 7\% and reducing forgetting by 3\% on LIBERO-LONG.
    Moreover, real-world deployment also demonstrates its effectiveness and efficiency.
\end{itemize}

\section{Related Works}
\label{sec:formatting}
\noindent\textbf{PEFT for Lifelong Learning.}
Parameter-efficient fine-tuning (PEFT) has exhibited strong performance in natural language processing~\cite{zhang2022continual,zhao2024sapt}
and computer vision~\cite{chen2022adaptformer,jia2022visual}, 
serving as a crucial way for continuous adaptation in lifelong learning. They mainly rely on prompt-based tuning~\cite{lester2021power}
to adapt a pre-trained model to new tasks. Despite the success of PEFT in lifelong learning, its application in robotics remains largely unexplored. To narrow the gap, a few works~\cite{schmied2023learning,lee2024incremental,hong2025hand} tend to explore it. TAIL~\cite{liu2023tail} learns task-specific adapters and relies on oracle task identifiers for knowledge retrieval. ABPFT~\cite{lu2024learning} considers the use of adapter technology through reinforcement learning. Instead, OMLA~\cite{zhu2025efficient} uses a meta-learning objective to learn a prior over the adapter. 
Our work is related to the PEFT-style adaptation paradigm while having a different focus. Rather than being a parameter-minimal adapter method, LiMoDE builds upon a TAIL and introduces additional dynamic and lifelong experts to explicitly model reusable skills and cross-task adaptation. This design trades moderate additional parameters for stronger lifelong adaptation, leading to a favorable performance-efficiency trade-off.

\begin{figure*}[t!]
\begin{center}
\includegraphics[width=0.9\linewidth]{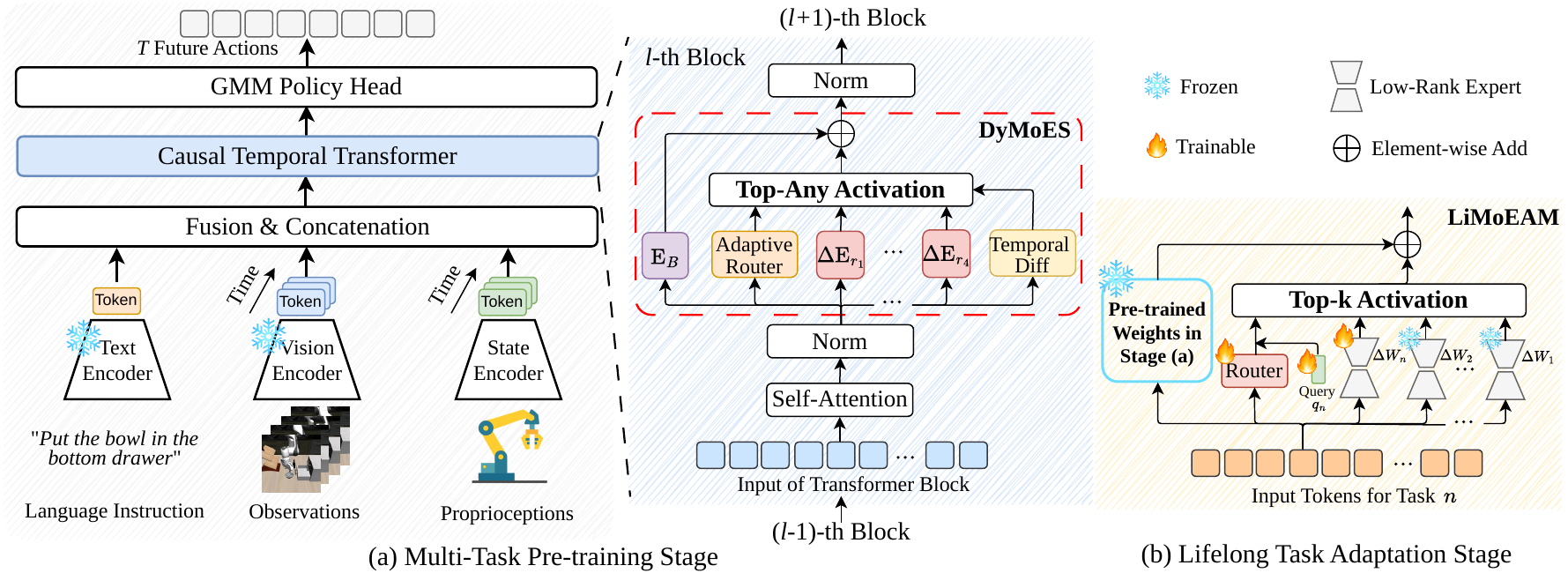}
\end{center}
\vspace{-10pt}
\caption{\textbf{Overview of the proposed Lifelong Mixture of Dynamic Experts (\textit{LiMoDE})}. It is a two-stage learning scheme that consists of a Dynamic Mixture of Experts Structure (DyMoES) and a Lifelong Mixture of Experts Adaptation Mechanism (LiMoEAM). In the pre-training stage, DyMoES activates a varied number of heterogeneous experts conditioned on motion information to learn skills in short-term manipulations. In the lifelong learning stage, LiMoEAM progressively learns new low-rank experts and combines them with previous ones to model task interaction for efficient adaptation. All weights obtained in the first stage are frozen in both adaptation and evaluation.}
\label{framework}
\vspace{-5pt}
\end{figure*}

\noindent \textbf{Mixture of Experts (MoE).}
It sparsely activates experts via a router network and has gained significant attention, inspiring a wide range of methodologies~\cite{dai2024deepseekmoe,liu2024deepseek,csordas2024moeut,csordas2024switchhead}. Follow-up works~\cite{nguyen2024statistical,wu2024routing,puigcerver2023sparse} further improve it by learning a finer input-conditioned router and performing an implicit soft assignment to each expert. In the context of large model fine-tuning, MoE is adopted to merge separate LoRAs for different tasks~\cite{liu2024adamole,wu2024mixture,wu2024omni}. In robot learning, some works~\cite{huang2025moe,wang2024sparse,reuss2024efficient,shen2025expertise} apply MoE to diffusion policy~\cite{chi2025diffusion}, enabling parameter-efficient scaling. \textit{This work explores the potential of MoE in lifelong robot learning by devising two novel structures, integrating heterogeneous experts, a motion-aware top-to-any router, and progressive expert learning. Their unique orchestration and task-specific adaptation constitute a significant contribution to the field.}

\noindent\textbf{Skill Learning for Robots.}
Skill learning is gaining increasing attention in acquiring new abilities and improving performance. Early approaches represent a skill by hand-crafted features~\cite{niekum2015learning}. With the development of deep learning, parameterization, latent codes, and MoE have become three strategies to learn skills from demonstrations. GSC~\cite{mishra2023generative} parameterizes several pre-defined skills as a set of variables, predicted by a diffusion policy, and SkillDiffuser~\cite{liang2024skilldiffuser} learns discrete skill codes that are highly correlated with language instructions. SDP~\cite{wang2024sparse} conceptualizes the experts in MoE as specialized skills and selectively activates them. Recently, PPL~\cite{yao2025think} utilizes reusable primitives for task adaptation. \textit{Differently, our framework learns base skills via the combination of a varied number of experts triggered by short-term motion during pre-training, and adaptively combines progressively learned experts with the current one during adaptation to enhance task interaction.}

\section{Preliminaries}
\noindent\textbf{Mixture of Experts.}
It comprises a router $R(\bm{x}):\mathbb{R}^{C}\rightarrow \mathbb{R}^{k}$ and $N_E$ expert networks $\{E_i(\bm{x})\}_{i=1}^{N_E}$,
where $\bm{x}\in \mathbb{R}^{d\times C}$ is the token embedding, $C$ is the input dimension, $k$ indicates the number of activated experts and $d$ is the token number. $R(\bm{x})$ selectively activates $k$ experts and MoE is formulated as:
{\setlength\abovedisplayskip{3pt}
\setlength\belowdisplayskip{3pt}
\begin{equation}\label{moe}
    \operatorname{MoE}(\bm{x}) = {\textstyle \sum_{i=1}^k} R(\bm{x})_i\cdot E_i(\bm{x}),
\end{equation}}where $\sum_{i}R(\bm{x})_i=1$ and $k$ controls the sparsity. We replace the FFN with the MoE layer in every other transformer block.  

Despite its effectiveness in multi-task robot learning~\cite{huang2025moe,wang2024sparse,reuss2024efficient,shen2025expertise}, it is still under-explored in lifelong robot learning.

\noindent\textbf{Lifelong Robot Learning.}
Assume we have a robot task set $\mathcal{T}=\{\mathcal{T}_i\}^{|\mathcal{T}|}_{i=1}$, where each demo $\mathcal{T}_i$ contains historical states ${\bm{s}}$, actions ${\bm{a}}$, and a language instruction ${\bm{g}}$ specifying the task.

\textit{1) Multi-Task Pre-Training.}
Its goal is to learn a policy network $\pi_{\theta}(\bar{\mathbf{a}}| {\mathbf{s}}, \mathbf{g})$, parameterized by $\theta$ and predicting $T$ future actions $\bar{\mathbf{a}} = \{\bar{\mathbf{a}}_{o+t}\}_{t=0}^{T-1}$ from the current time step $o$, where each action $\bar{\mathbf{a}}_i=[\Delta x, \Delta y, \Delta z, \Delta A_x, \Delta A_y, \Delta A_z, gripper]$ consists of relative translation ($\Delta \{x,y,z\}$), Euler-angle-based relative rotation $\Delta A_{\{x,y,z\}}$, and the gripper state $gripper\in\{0, 1\}$. The standard cloning loss~\cite{bain1995framework} is used over $\mathcal{T}$:
{\setlength\abovedisplayskip{3pt}
\setlength\belowdisplayskip{3pt}
\begin{equation}
\mathcal{L}_{BC} = {\textstyle \sum_{i=1}^{|\mathcal{T}|}}\mathbb{E}_{s_t,a_t \sim \mathcal{T}_i} \left[ {\textstyle \sum_{t=0}^{l_i}} \mathcal{L}( \pi_{\theta}(\bar{\mathbf{a}}_t|\mathbf{s}_t, \mathbf{g}_i), \mathbf{a}_t) \right],
\label{bcloss}
\end{equation}}where $\mathcal{L}$ is a negative log-likelihood loss and $l_i$ is the length of demonstrations for task $\mathcal{T}_i$.

\textit{2) Lifelong Learning Stage.}
This stage builds upon the skills acquired during multi-task pre-training, and the objective becomes to incrementally learn over a stream of tasks while retaining performance on old ones. The pre-trained policy continues to encounter a sequence of $N$ tasks, denoted as $\{\mathcal{T}_{|\mathcal{T}|+n}\}_{n=1}^N$, and each task contains demonstrations $\mathcal{T}_{|\mathcal{T}|+n} = \{\tau_{n}^i\}_{i=1}^{|\mathcal{T}_{|\mathcal{T}|+n}|}$ for learning. Here, $\mathcal{D}_k$ only contains data from the current task, and $s_t$ should be interpreted as $s_{\leq t}$.

\section{Methodology}
\textbf{Overview.}
To explicitly model the cross-task interaction mentioned in Sec.~\ref{intro}, we design two novel modules: a Dynamic MoE Structure (DyMoES) and a Lifelong MoE Adaptation Mechanism (LiMoEAM). DyMoES formulates base skills via MoE, activating a varied number of experts based on short-term motion, which novelly links motion to skill learning and is absent in the literature. In the adaptation stage, LiMoEAM progressively learns new experts. Combining them with previous ones models the new-to-new adaptation, while additionally integrating base skills leads to the base-to-new adaptation. 
Fig.~\ref{framework} details the structure of the \textit{\textbf{LiMoDE}}. Inputs are respectively encoded by the CLIP~\cite{radford2021learning} and an MLP, and further projected into a joint embedding space, where FiLM~\cite{perez2018film} fuses vision and language tokens. The fused embedding is concatenated with proprioception tokens and sent to a causal transformer. Finally, an action head predicts the distribution.

\subsection{Multi-Task Pre-Training Stage}\label{MTPT}
Base tasks share skills within short-term actions that can be adapted to similar actions in new tasks. We learn a set of reusable skills with shared knowledge across tasks via a Dynamic Mixture of Experts Structure (DyMoES). It has two complementary components: heterogeneous experts that learn a set of reusable low-rank experts with different capacities and a visual-dynamics-conditioned router that determines which experts to activate. The architecture specifies \emph{what capacity is available}, and the gating specifies \emph{which capacity is used}.

\noindent\textbf{Dynamic Mixture of Expert Structure (DyMoES).} Previous works~\cite{huang2025moe,wang2024sparse} typically implement experts as feedforward network layers, which account for the majority of parameters. To reduce trainable parameters, inspired by~\cite{wu2024mixture}, we decompose an expert into two parts and propose the mixture of low-rank experts. 
Let $E_i(\cdot)$ be the $i$-th expert and we reformulate it as the sum of a base expert $E_{B}(\cdot)$, and a low-rank expert $\Delta E_i(\cdot)$,~\textit{i.e.,} $E_i(\bm{x}) = E_{B}(\bm{x}) + \Delta E_i(\bm{x})$.

\textit{1) Heterogeneous Experts.} $\Delta E_i(\cdot)$ is homogeneous, and varying complexity of skills in short-term motions necessitates experts with diverse capabilities. We introduce low-rank expert $\Delta E_{r_i}(\cdot)=\bm{A}_{r_i}\bm{B}_{r_i}$, where $A_i$ and $B_i$ are two low-rank matrix of rank $r_i$ that controls the capacity, This gives heterogeneous experts $E_i(\bm{x}) = E_{B}(\bm{x}) + \Delta E_{r_i}(\bm{x})$ and Eq.~(\ref{moe}) becomes:
{\setlength\abovedisplayskip{3pt}
\setlength\belowdisplayskip{3pt}
\begin{equation}\label{MoHE}
    \operatorname{MoE}(\bm{x}) = E_{B}(\bm{x}) + {\textstyle \sum_{i=1}^k} R(\bm{x})_i\cdot \Delta E_{r_i}(\bm{x}),
\end{equation}}where $E_{B}(\bm{x})$ serves as a shared expert in MoE and Eq.~(\ref{MoHE}) enables more specialized experts for varied complexities.

\textit{2) Visual-dynamics-conditioned Router.} Router in Eq.~(\ref{moe}) is $R(\bm{x}) = \text{softmax}(\bm{W}_g^{T} \bm{x})$ with gating network $\bm{W}_g \in \mathbb{R}^{C \times k}$. It requires tuning $k$ for optimal performance, and activating $k$ experts is insufficient for skills with varying difficulties. 

Note that the intensity of short-term motions varies. 
Smooth phases, such as reaching or transport, are usually easier to model, while visually dynamic or contact-rich phases, such as grasping, placing, opening, or releasing, often involve more complex and less smooth action distributions. Since explicit phase labels are unavailable, visual temporal differences serve as a lightweight proxy for local visual dynamics and phase-level difficulty, which is empirically demonstrated in our experimental findings in Fig. \ref{vis1}. We thus design a novel router to adaptively decide the number of activated experts based on the short-term motion. Given a gating network $\bm{W}_{\text{router}} \in \mathbb{R}^{C \times N_{E}}$, where $i$-th row acts as the query for $i$-th expert, a query score for each expert is measured: $s(\bm{x}) = \text{Cosine}(\bm{x}, \bm{W}_{\text{router}})$. An activation indicator $g(\bm{x}) = \text{sign} \left( \sigma \left( s(\bm{x}) \right) - [1 - \sigma(\bm{M}(\bm{x}))] \right)$ is computed, where $\sigma$ and $\text{sign}(\cdot)$ refer to the sigmoid and Sign function. $\bm{M}$ is a visual-dynamics-conditioned threshold from the temporal difference between current and historical visual tokens: $\bm{M} = \text{MLP}(f^{vis}_{o}-f^{vis}_{o-T})$, where $f^{vis}_{o}$ denotes the current visual token. 
It provides a learnable cue for the router to infer when additional expert capacity may be useful. The DyMoES layer is formulated by:
{\setlength\abovedisplayskip{3pt}
\setlength\belowdisplayskip{3pt}
\begin{equation}\label{dymohe}
\scalebox{1.01}{$
     \operatorname{DyMoES}(\bm{x}) = E_{B}(\bm{x}) + \sum_{g(\bm{x})_i>0} a(\bm{x})_i \Delta E_{r_i}(\bm{x}),  $}
\end{equation}}where $\sum_i g(\bm{x})_i=K\in[0, N_E]$, called the top-any activation, and $a(\bm{x})_i=\operatorname{Softmax}\left(g(\bm{x})_i\right)$. Intuitively, it activates more experts for skills with larger visual dynamics. A larger learned temporal-difference score decreases the activation threshold $1 - \sigma(\bm{M}(\bm{x}))$, making additional experts easier to activate. This design does not assume that raw motion magnitude alone determines task difficulty. Instead, it allows the router to adaptively allocate expert capacity to temporally distinct manipulation phases. We copy the gradient of $g(\bm{x})$ to $\sigma \left( s(\bm{x}) \right) - [1 - \sigma(\bm{M}(\bm{x}))]$ to bypass the Sign function.

\textit{3) Router-Decorrelation Regularizer (RDR).}
Works~\cite{wang2024sparse,reuss2024efficient} leverage load balancing loss to eliminate unbalanced routing distributions in homogeneous MoE. However, they are not applicable in a heterogeneous case, where the number of activated experts is varied, and all experts might be activated. 

To constrain load balancing and the number of activated experts, we adopt a lightweight router-decorrelation regularizer $\mathcal{L}_{\text{RDR}}$ on the Gramian matrix of the gating network $\bm{W}_{\text{router}}$:
{\setlength\abovedisplayskip{2pt}
\setlength\belowdisplayskip{2pt}
\begin{equation}\label{sparseloss}
    \mathcal{L}_{\text{RDR}} = \frac{1}{2}\|\bm{W}_{\text{router}}^{T}\bm{W}_{\text{router}} - \bm{I}_{N_E}\|_2^2,
\end{equation}}where $\bm{I}_{N_E}$ is an identity matrix with dimension $N_E$. Note that orthogonality- and decorrelation-inspired routing regularizers have been explored in MoE to improve routing diversity and reduce redundant expert assignment \cite{shazeer2017outrageously,lepikhin2020gshard,fedus2022switch}. Recent works regularize router weights or routing decisions to encourage more diverse and discriminative expert selection \cite{omi2025load,guo2026advancing,hu2026synergistic} for vanilla MoE. Different from these general MoE regularizers, our regularizer is used as a lightweight router-decorrelation term within a visual-dynamics-conditioned routing mechanism. It should be interpreted as an auxiliary router regularizer rather than a direct expert-dissimilarity loss, encouraging diverse routing queries and complementing the proposed visual-dynamics-conditioned top-any activation. Moreover, it also helps avoid activating all experts simultaneously and promotes the sparsity of experts, improving efficiency. 

\noindent\textbf{Disentangling Architecture and Gating.} DyMoES contains two complementary but distinct components. 
The heterogeneous expert bank is an architectural design that determines \emph{what adaptation capacity is available} by providing low-rank experts with different ranks. 
The visual-dynamics-conditioned top-any gate is a routing design that determines \emph{which capacity is used} by activating an input-dependent number of experts. 
Therefore, the empirical gains should not be attributed solely to increased parameters or solely to the gating rule. 
In Sec.~\ref{abl}, we explicitly disentangle these factors through ablations on MoE, dynamics, heterogeneity, shared thresholds, Gumbel-Sigmoid routing, and homogeneous-rank experts.


\begin{figure}[t]
  \centering
  \includegraphics[width=0.65\linewidth]{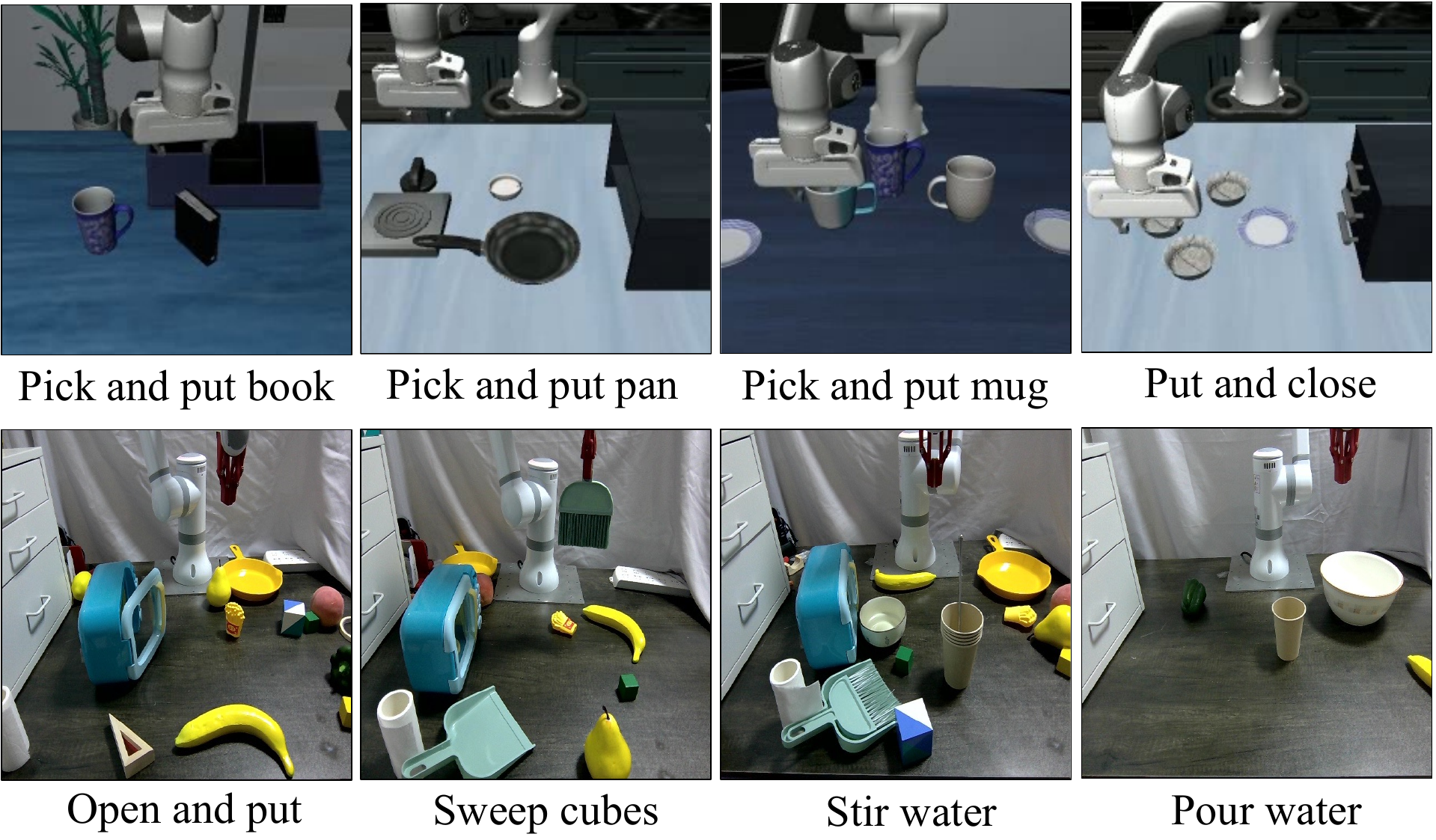}
  \vspace{-4pt}
   \caption{\textbf{Experimental setup}. \textbf{Top:} LIBERO. \textbf{Bottom}: Real world.}
   \label{dataset}
   \vspace{-10pt}
\end{figure}
\subsection{Lifelong Task Learning Stage}\label{LFTL}
\noindent\textbf{Lifelong MoE Adaptation Mechanism (LiMoEAM).} After learning reusable skills, the lifelong learning stage adapts these skills and new skills to newly arrived tasks. Thus, the LiMoEAM is introduced to incrementally build a library of experts and dynamically assemble them for adaptation.

\textit{1) Mixture of Lifelong Experts.}
Previous works learn task-specific adapters during adaptation, which need an identifier for task retrieval. Such a paradigm lacks flexibility and hinders knowledge sharing across tasks. Since low-rank experts capture task-specific knowledge from a variety of tasks, they can be complementary and leveraged to solve a new task. Following this intuition, we progressively learn and combine low-rank experts from prior tasks $j\leq n-1$ for the $n$-th task. This is why we call them lifelong experts.

Specifically, assume we have learned low-rank experts $\{\Delta E^{\text{life}}_j(\cdot)\}_{j=1}^{n-1}$ and a router network $\bm{W}_{j\leq n-1}\in\mathbb{R}^{C \times (n-1)}$. To learn the $n$-th task, a new low-rank expert $\Delta E^{\text{life}}_n(\cdot)$ and a query vector $\bm{q}_n\in\mathbb{R}^{C \times 1}$ is initiated. Then $\bm{W}_{j\leq n-1}$ and $\bm{q}_n$ are concatenated to form a new router $\bm{W}_{j\leq n}$$=[\bm{W}_{j\leq n-1}; \bm{q}_n]\in\mathbb{R}^{C \times n}$. The expert combination coefficient is then obtained $C(\bm{x}) = \operatorname{softmax}\left(\operatorname{Top-k}(\bm{W}^T_{j\leq n} \bm{x})\right)$, where $\operatorname{Top-k}(\cdot)$ retains the top-$k$ value while setting all others to zero. Finally, the skill for input $\bm{x}$ is represented by:
{\setlength\abovedisplayskip{3pt}
\setlength\belowdisplayskip{3pt}
\begin{equation}\label{pmoe}
    \operatorname{LiMoEAM}(\bm{x}) = W^T_{\text{pretrained}}\bm{x} + {\textstyle \sum_{j=1}^k}  C(x)_j \Delta E^{\text{life}}_j(\bm{x}),
\end{equation}}For the next new task, we freeze $\{\Delta E^{\text{life}}_j(\cdot)\}_{j=1}^{n}$ as acquired knowledge and initialize another learnable expert and the query vector. This process ends after learning all the new tasks.

\textit{2) Replay Strategy.}
Parameters of the router are continuously updated during adaptation, and directly using them may cause incorrect expert selection and task forgetting. We store their coefficients for supervision during learning the $n$-th new task.

During adaptation, we store router input context and its ouptut activation for each task as $\mathcal{M}=\{\left(\bm{e}_{i,j}, C_j(\bm{e_{i,j}})\right)\}_{j<n}$, and enforce the consistency by requiring the router to generate coefficients that closely match the stored historical coefficients, given the same context.
We omit the subscript of the layer for clarification. Then the constraint is given as:
{\setlength\abovedisplayskip{3pt}
\setlength\belowdisplayskip{3pt}
\begin{equation}\label{replayloss}
\scalebox{0.95}{$
\mathcal{L}_{\text{RS}} = \mathbb{E}_{\left(\bm{e}_{i,j}, C(\bm{e_{i,j}})\right)\sim\mathcal{M}}\frac{1}{2}\|C_n(\bm{e}_{i,j})_{1:j} -  C_j(\bm{e_{i,j}})\|_2^2.$}
\end{equation}}Although the replay strategy is widely used in robotic lifelong learning \cite{chaudhry2019continual,xie2022lifelong,zhao2024sapt}, they usually require storing entire trajectories and additional forward propagation to process them. Our strategy differs in that it supervises expert-composition coefficients to mitigate router drift using low-dimensional embeddings and coefficients, rather than replaying raw trajectories for optimization. Compared to conventional ER methods, our strategy substantially reduces storage from 1.4 GB/task to 60.7 MB/task, while avoiding replay-time forward passes through the full policy, resulting in significant reductions in both computational overhead and storage requirements.

\begin{table*}[ht!]
\caption{\textbf{Lifelong learning performance on the LIBERO benchmark}. The best values are highlighted in bold, and the dash (--) indicates no results reported. The reported values are averages over three seeds, and all metrics are measured based on success rates (\%). The upward ($\uparrow$) and downward ($\downarrow$) indicate higher values and lower values are better, respectively. $\dagger$ refers to re-implementation.
}\label{liberosuites}
\centering
\resizebox{0.8\textwidth}{!}{%
\begin{tabular}{@{}@{\extracolsep{\fill}}c|ccc|ccc|ccc|ccc@{}}
\toprule
\multirow{2}{*}{Method} & \multicolumn{3}{c}{LIBERO-OBJECT} & \multicolumn{3}{c}{LIBERO-GOAL} & \multicolumn{3}{c}{LIBERO-SPATIAL} & \multicolumn{3}{c}{LIBERO-LONG}\\
\cmidrule(lr){2-4} \cmidrule(lr){5-7} \cmidrule(lr){8-10} \cmidrule(lr){11-13}
 & FWT ($\uparrow$) & BWT ($\downarrow$) & AUC ($\uparrow$) & FWT ($\uparrow$) & BWT ($\downarrow$) & AUC ($\uparrow$) & FWT ($\uparrow$) & BWT ($\downarrow$) & AUC ($\uparrow$) & FWT ($\uparrow$) & BWT ($\downarrow$) & AUC ($\uparrow$) \\
\midrule
SeqFT~\cite{liu2023libero} & 62.0 $\pm$ 1.0 & 63.0 $\pm$ 2.0 & 30.0 $\pm$ 1.0 & 55.0 $\pm$ 1.0 & 70.0 $\pm$ 1.0 & 23.0 $\pm$ 1.0 & 72.0 $\pm$ 1.0 & 81.0 $\pm$ 2.0 & 20.0 $\pm$ 1.0 & 48.0 $\pm$ 2.0 & 64.0 $\pm$ 2.0 & 12.0 $\pm$ 1.0 \\
ER~\cite{chaudhry2019continual} & 56.0 $\pm$ 1.0 & 24.0 $\pm$ 1.0 & 49.0 $\pm$ 1.0 & 53.0 $\pm$ 1.0 & 36.0 $\pm$ 1.0 & 47.0 $\pm$ 2.0 & 65.0 $\pm$ 3.0 & 27.0 $\pm$ 3.0 & 56.0 $\pm$ 1.0 & 47.0 $\pm$ 2.0 & 22.0 $\pm$ 1.0 & 53.0 $\pm$ 1.0 \\
L2M~\cite{schmied2023learning} & 16.0 $\pm$ 4.0 & 14.0 $\pm$ 5.0 & 8.0 $\pm$ 2.0 & 12.0 $\pm$ 2.0 & 28.0 $\pm$ 1.0 & 7.0 $\pm$ 1.0 & 6.0 $\pm$ 1.0 & 24.0 $\pm$ 2.0 & 4.0 $\pm$ 1.0 & 6.0 $\pm$ 1.0 & 24.0 $\pm$ 1.0 & 5.0 $\pm$ 1.0 \\
TAIL~\cite{liu2023tail} & 58.0 $\pm$ 3.0 & 50.0 $\pm$ 3.0 & 70.0 $\pm$ 4.0 & 54.0 $\pm$ 1.0 & 66.0 $\pm$ 2.0 & 67.0 $\pm$ 2.0 & 51.0 $\pm$ 1.0 & 56.0 $\pm$ 2.0 & 60.0 $\pm$ 1.0 & 45.0 $\pm$ 0.0 & 61.0 $\pm$ 3.0 & 54.0 $\pm$ 2.0 \\
IsCiL~\cite{lee2024incremental} & 57.0 $\pm$ 5.0 & 6.0 $\pm$ 2.0 & 66.0 $\pm$ 4.0 & 54.0 $\pm$ 2.0 & 32.0 $\pm$ 2.0  & 60.0 $\pm$ 3.0 & 38.0 $\pm$ 2.0 & 23.0 $\pm$ 2.0 & 44.0 $\pm$ 3.0 & 44.0 $\pm$ 2.0 & 33.0 $\pm$ 4.0 & 32.0 $\pm$ 2.0 \\
LOTUS~\cite{wan2024lotus} & 74.0 $\pm$ 3.0 & 11.0 $\pm$ 1.0 & 65.0 $\pm$ 3.0 & 61.0 $\pm$ 3.0 & 30.0 $\pm$ 1.0 & 56.0 $\pm$ 1.0 & - & - & - & 29.0 $\pm$ 3.0 & 17.0 $\pm$ 1.0 & 30.0 $\pm$ 2.0 \\
BUDS~\cite{zhu2022bottom} & 52.0 $\pm$ 2.0 & 21.0 $\pm$ 1.0 & 47.0 $\pm$ 1.0 & 50.0 $\pm$ 1.0 & 39.0 $\pm$ 1.0 & 42.0 $\pm$ 1.0 & - & - & - & - & - & - \\
M2Distill~\cite{roy2025m2distill}$^\dagger$ & 75.0 $\pm$ 3.0 & 8.0 $\pm$ 5.0 & 69.0 $\pm$ 4.0 & \textbf{71.0 $\pm$ 1.0} & 20.0 $\pm$ 3.0 & 57.0 $\pm$ 2.0 & 74.0 $\pm$ 1.0 & \textbf{11.0 $\pm$ 1.0} & 61.0 $\pm$ 2.0 & 46.0 $\pm$ 3.0 & 20.0 $\pm$ 2.0 & 47.0 $\pm$ 3.0 \\
PPL~\cite{yao2025think}$^\dagger$ & 77.0 $\pm$ 2.0 & 9.0 $\pm$ 2.0 & 72.0 $\pm$ 2.0 & 68.0 $\pm$ 2.0 & 22.0 $\pm$ 2.0 & 63.0 $\pm$ 2.0 & 70.0 $\pm$ 2.0 & 17.0 $\pm$ 2.0 & 64.0 $\pm$ 2.0 & 48.0 $\pm$ 3.0 & 23.0 $\pm$ 2.0 & 49.0 $\pm$ 3.0 \\
\midrule 
\cellcolor{pink!20}\textbf{Ours} & \cellcolor{pink!20}\textbf{81.0 $\pm$ 2.0} & \cellcolor{pink!20}\textbf{5.0 $\pm$ 1.0} & \cellcolor{pink!20}\textbf{76.0 $\pm$ 2.0} & \cellcolor{pink!20}70.0 $\pm$ 2.0 & \cellcolor{pink!20}\textbf{18.0 $\pm$ 1.0} & \cellcolor{pink!20}\textbf{69.0 $\pm$ 3.0} & \cellcolor{pink!20}\textbf{75.0 $\pm$ 3.0} & \cellcolor{pink!20}14.0 $\pm$ 1.0 & \cellcolor{pink!20}\textbf{69.0 $\pm$ 3.0} & \cellcolor{pink!20}\textbf{55.0 $\pm$ 3.0} & \cellcolor{pink!20}\textbf{14.0 $\pm$ 1.0} & \cellcolor{pink!20}\textbf{58.0 $\pm$ 3.0} \\
\bottomrule
\end{tabular}}
\end{table*}

\begin{table}
\caption{\textbf{Performance on real-world tasks.} Success rate (\%) over 20 trials is reported. Our \textbf{\textit{LiMoDE}} performs best in both stages.}
\label{realword}
\centering
\resizebox{0.85\linewidth}{!}{
\begin{tabular}{cccc}
\toprule
\multirow{2}{*}{Task} & \multicolumn{3}{c}{Methods} \\
\cline{2-4}
 & SDP~\cite{wang2024sparse} & OpenVLA~\cite{kim2024openvla} &  \textbf{\textit{LiMoDE}} (Ours) \\
\midrule
& \multicolumn{3}{c}{Multi-Task Fine-tuning} \\
\midrule
Pretrain Task 1 & 60.0\% & 55.0\% & \cellcolor{pink!20}\textbf{70.0\%} \\
Pretrain Task 2 & 20.0\% & 15.0\% & \cellcolor{pink!20}\textbf{45.0\%} \\
Pretrain Task 3 & 30.0\% & 20.0\% & \cellcolor{pink!20}\textbf{65.0\%} \\
\midrule
& \multicolumn{3}{c}{Lifelong Learning} \\
\midrule
Task & SequFT~\cite{liu2023libero} & PPL~\cite{yao2025think} & \textbf{\textit{LiMoDE}} (Ours) \\
\midrule
Lifelong Task 1 & 40.0\% & 55.0\% & \cellcolor{pink!20}\textbf{65.0\%} \\
Lifelong Task 2 & 20.0\% & 35.0\% & \cellcolor{pink!20}\textbf{55.0\%} \\
Lifelong Task 3 & 15.0\% & 30.0\% & \cellcolor{pink!20}\textbf{60.0\%} \\
Lifelong Task 4 & 10.0\% & 25.0\% & \cellcolor{pink!20}\textbf{50.0\%} \\
Lifelong Task 5 & 10.0\% & 20.0\% & \cellcolor{pink!20}\textbf{40.0\%}\\
\bottomrule
\vspace{-20pt}
\end{tabular}}
\end{table}


\section{Experiments}
We conduct comprehensive experiments on \textbf{\textit{LiMoDE}} and answer: 1) In Sec.~\ref{exp1} and \ref{exp2}, how well does it perform? 2) In Sec. \ref{abl}, how does key designs impact performance? 3) In Sec. \ref{visu}, how does it work in real-world scenarios?

\subsection{Experimental Settings}
\noindent\textbf{Simulation.}
Our extensive evaluations are conducted on the LIBERO benchmark~\cite{liu2023libero}, as shown in Fig.~\ref{dataset}. It consists of a 6-DOF robot arm equipped with a parallel gripper operating in a tabletop environment. We systematically test on four task suites: Goal, Spatial, Object, and Long, where 10 manipulation tasks arrive sequentially in each suite to challenge the robot's ability to transfer knowledge related to task goals, spatial information, and objects across different domains. Following~\cite{liu2023tail,yao2025think}, we leverage the LIBERO-90 for large-scale pre-training. With the formulation in Sec.~\ref{MTPT}, the robot is required to generate continuous actions driven by multi-modal inputs.

\noindent\textbf{Real-world Tasks.}
We design 8 tasks with 90 trajectories each via a 6-DoF robot arm. Tasks contain not only simple ``pick \& place'' but also dextrous tasks like ``pour water'' and more difficult tasks that need multi-step operations. Three tasks that contain necessary skills for lifelong tasks are used for pre-training, with the remaining for lifelong learning. \textit{Pretrain tasks}: ``Pick \& place lemon'', ``Open microwave \& put chips'', ``Pour water from cup to bowl''. \textit{Lifelong tasks}: ``Pick \& place banana'', ``Stir water'', ``Sweep cubes'', ``Stack cubes'', ``Pick \& place all cubes''. Note that we exclude tasks like ``fold clothes'' that require specialized tactile sensing or physical modeling, which are in a distinct domain from our setting. 

\noindent\textbf{Evaluation Metrics.}
Following~\cite{liu2023libero,yao2025think}, we use forward transfer (FWT) computed by the maximum success rate (SR), negative backward transfer (NBT) measuring increases in average SR on previous tasks, and area under the success rate curve (AUC) as lifelong learning metrics. A higher FWT  indicates faster adaptation, and a lower NBT means less forgetting of old tasks. Finally, a higher AUC means an overall better performance considering both FWT and NBT.

\noindent\textbf{Implementation Details.}
\textbf{\textit{LiMoDE}} is built on TAIL~\cite{liu2023tail}. In the first stage, we adopt a similar training strategy as in~\cite{liu2023tail} to pretrain \textbf{\textit{LiMoDE}} on the LIBERO-90, where $T=10$, $N_E=4$, and the rank $r_i=2^{i+1}$. In the second stage, we freeze all pre-trained weights and $k=2$. The model is trained for 10 epochs on each new task and evaluated every 2 epochs, with the AdamW optimizer and a learning rate of $10^{-4}$. The batch size is 32, and images from the static camera are resized. We perform expert coefficient replay for 10 epochs after each task, and combine the top-2 experts. For real-world evaluation, we also use images from the static camera and train the model for 20 epochs. Results are averaged over 20 trials. To make fair comparisons, GMM head~\cite{liu2023tail} is adopted as the policy head. During training, we optimize the sum of $\mathcal{L}_{\text{BC}}$ and $\mathcal{L}_{\text{RDR}}$ over all DyMoES in multi-task pre-training, and additionally include $\mathcal{L}_{\text{RS}}$ for the lifelong learning stage. In inference, following~\cite{liu2023libero}, the mean of the Gaussian distribution with the highest density is used for robot execution.


\begin{table}
\caption{\textbf{Auxiliary evaluation of multi-task performance (\%) on LIBERO}. Zero standard deviation means no average performance. $\dagger$ denotes re-implementation with only LIBERO-90 pre-training.}\vspace{-5pt}
  \label{liberomultitasks}
  \begin{center}
  \resizebox{0.92\linewidth}{!}{
  \begin{tabular}{@{}@{\extracolsep{\fill}}c|cccc|c@{}}
    \toprule
    Method & Spatial & Object & Goal & Long & Average \\
    \midrule
    DP~\cite{chi2025diffusion} & 78.3$\pm$0.0 & 92.5$\pm$0.0 & 68.3$\pm$0.0 & 50.5$\pm$0.0 & 72.4$\pm$0.0 \\
    SDP~\cite{wang2024sparse} & 78.5$\pm$0.0 & 87.5$\pm$0.0 & 73.5$\pm$0.0 & 64.8$\pm$0.0 & 75.1$\pm$0.0 \\
    OpenVLA~\cite{kim2024openvla} & 84.7$\pm$0.9 & 88.4$\pm$0.8 & 79.2$\pm$1.0 & 53.7$\pm$1.3 & 76.5$\pm$0.6 \\
    UniAction~\cite{zheng2025universal} & 65.0$\pm$0.0 & 78.0$\pm$0.0 & 68.0$\pm$0.0 & 47.0$\pm$0.0 & 64.5$\pm$0.0 \\
    UniVLA~\cite{bu2025learning}$^\dagger$ & \textbf{92.6$\pm$0.0} & 93.8$\pm$0.0 & 86.6$\pm$0.0 & 63.0$\pm$0.0 & 84.0$\pm$0.0 \\
    PPL~\cite{yao2025think}$^\dagger$ & 85.0$\pm$3.0 & 86.0$\pm$2.0 & 86.0$\pm$1.0 & 80.0$\pm$3.0 & 84.0$\pm$2.0 \\
    \midrule
    \cellcolor{pink!20}\textbf{Ours} & \cellcolor{pink!20}90.2$\pm$1.0 & \cellcolor{pink!20}\textbf{94.2$\pm$0.5} & \cellcolor{pink!20}89.2$\pm$0.7 & \cellcolor{pink!20}\textbf{81.4$\pm$1.0} & \cellcolor{pink!20}\textbf{88.7$\pm$0.6} \\
    \bottomrule
  \end{tabular}}
  \end{center}
  \vspace{-15pt}
\end{table}

\subsection{Performance on Lifelong Learning}\label{exp2}
\noindent\textbf{Results in Simulation.}
We compare our \textbf{\textit{LiMoDE}} with the following lifelong baselines: Experience Replay (ER)~\cite{chaudhry2019continual}, BUDS~\cite{zhu2022bottom}, Sequential fine-tuning (SeqFT)~\cite{liu2023libero}, L2M~\cite{schmied2023learning}, TAIL~\cite{liu2023tail}, LoTuS~\cite{wan2024lotus}, and IsCiL~\cite{lee2024incremental}, and recent M2Distill~\cite{roy2025m2distill} and PPL~\cite{yao2025think}. In Tab.~\ref{liberosuites}, we conduct comparisons on the LIBERO benchmark, where \textbf{\textit{LiMoDE}} achieves top-tier performance across all suites. On LIBERO-OBJECT, our approach gives the highest FWT and AUC scores while maintaining the lowest BWT, significantly outperforming strong recent baselines such as M2Distill and PPL. In LIBERO-GOAL, we obtain a competitive FWT, the best BWT, and AUC, where only M2Distill shows a slightly better FWT.
This indicates that while M2Distill is effective in reducing representation drift through multi-modal distillation, \textbf{\textit{LiMoDE}} achieves a better overall balance between FWT and forgetting mitigation.
For the more challenging LIBERO-SPATIAL, our method delivers outstanding performance with $75.0 \pm 3.0$ FWT and $69.0 \pm 3.0$ AUC, matching or exceeding all comparators.
Although M2Distill obtains a lower BWT, our method achieves stronger FWT and AUC, suggesting better overall lifelong adaptation.
On the demanding LIBERO-LONG, we achieve the highest scores across all metrics, demonstrating exceptional capability in handling long-horizon tasks that several baselines either failed or showed substantially inferior results.
In particular, M2Distill is distillation-based and does not explicitly model long-horizon skill composition, while PPL relies on primitive prompts but lacks explicit lifelong reuse and coefficient replay. In contrast, \textbf{\textit{LiMoDE}} dynamically reuses and composes experts across tasks, leading to clear advantages on long-horizon lifelong manipulation. These results show \textbf{\textit{LiMoDE}}'s outstanding performance, validating its effectiveness and robustness in diverse lifelong learning scenarios.

\begin{table}[!t]
\caption{\textbf{Ablation study} on (a) Key components and (b) design choices. Results on MTFT and lifelong learning are reported. Shared MoE, $M$, and $\Delta W_n$ refer to applying a shared combination of heterogeneous MoE to all tokens, learning a parameter $M$ in the router, and adaptation with only one expert, respectively. Frozen Router means that the router in LiMoEAM is frozen during adaptation.}
\label{ablation}
\centering
\resizebox{0.9\linewidth}{!}{ 
\begin{tabular}{llccccccc}
\toprule
& \multirow{2}{*}{Components} & \multirow{2}{*}{MTFT} & \multicolumn{3}{c}{Lifelong Learning (FWT)} \\
\cmidrule(lr){4-6}
 & & & OBJECT & GOAL & SPATIAL  \\
\midrule
\multirow{6}{*}{DyMoES} & Baselines & 64.3 & 58.0 & 54.0 & 51.0  \\
& + MoE & 69.2 & 64.0 & 58.0 &  57.0\\
& + Dynamics & 73.0 & 66.0 & 60.0 & 62.0 \\
& + Heterogeneity & 78.1 & 70.0 & 62.0 & 63.0 \\
& + $\mathcal{L}_{\text{RDR}}$ & 80.4 & 72.0 & 64.0 & 66.0\\
\midrule
\multirow{2}{*}{LiMoEAM} & + LiMoEAM & 85.6 & 77.0 & 67.0 & 71.0 \\
& + $\mathcal{L_{\text{RS}}}$ & \textbf{88.7} & \textbf{81.0} & \textbf{70.0} & \textbf{75.0} \\
\midrule
\multicolumn{6}{c}{(a) Study on key components in DyMoES and LiMoEA.}\\
\midrule
\multirow{5}{*}{DyMoES}    & Shared MoE & 86.2 & 78.0 & 67.0 & 71.0 \\
& Shared $\bm{M}$           & 87.4 & 79.0 & 68.0 & 73.0  \\
& Gumbel-Sigmoid & 86.7 & 77.0 & 67.0 & 72.0  \\
\cmidrule(lr){2-6}
& Homogeneity ($r_i$=8) & 86.4 & 78.0 & 67.0 & 72.0 \\
& Homogeneity ($r_i$=16) & 87.8 & 79.0 & 68.0 & 74.0 \\
\midrule
\multirow{3}{*}{LiMoEAM} & Frozen Router & 82.5 & 73.0 & 65.0 & 68.0 \\
& Shared $\Delta W_n$  & 83.8 & 75.0 & 70.0 & 70.0 \\
& Ours   & \textbf{88.7} & \textbf{81.0} & \textbf{70.0} & \textbf{75.0} \\
\midrule
\multicolumn{6}{c}{(b) Study on design choices in DyMoES and LiMoEAM.}\\
\vspace{-25pt}
\end{tabular}}
\end{table}


\noindent\textbf{Real-world Results.}
For MTFT, the MoE-based SDP~\cite{wang2024sparse} and the representative OpenVLA~\cite{kim2024openvla} are compared. For lifelong learning, the SeqFT~\cite{liu2023libero} and the recent primitive-learning approach PPL~\cite{yao2025think} are considered. We follow~\cite{yao2025think} to perform the evaluation. 
The results on real-world tasks are reported in Tab.~\ref{realword}, where our \textbf{\textit{LiMoDE}} consistently outperforms baselines. In the fine-tuning phase, \textbf{\textit{LiMoDE}} achieves over $2\times$ improvements for challenging tasks, demonstrating the effectiveness of learning skills by combining a varied number of heterogeneous experts. More importantly, through performing cross-task interaction during task adaptation, \textbf{\textit{LiMoDE}} maintains this advantage, and the consistent performance gaps demonstrate \textbf{\textit{LiMoDE}}'s capability in retaining previously acquired knowledge while effectively adapting to new tasks. These results validate the effectiveness of \textbf{\textit{LiMoDE}} in practical robot applications, highlighting its robust performance in both initial skill acquisition and continual task adaptation scenarios.

\subsection{Auxiliary Evaluation of Multi-Task Fine-Tuning}\label{exp1}
To complement the lifelong learning results in Sec. \ref{exp2}, we additionally verify whether the pre-training stage contributes to competitive multi-task performance. We select commonly used baselines, such as Diffusion Policy (DP)~\cite{chi2025diffusion}, OpenVLA~\cite{kim2024openvla}, SDP~\cite{wang2024sparse}, and UniAction~\cite{zheng2025universal}, as well as recent UniVLA~\cite{bu2025learning} and PPL~\cite{yao2025think} re-implemented under our setting for comparison.

The pre-trained \textbf{\textit{LiMoDE}} is evaluated on the remaining suites of LIBERO in Tab.~\ref{liberomultitasks} and achieves strong performance among the listed baselines.
While UniVLA shows strong performance in spatial reasoning, our method demonstrates better overall capability with the highest success rate in object generalization, goal adaptation, and long-horizon planning. In LIBERO-Long, it outperforms listed baselines by a considerable margin, while maintaining competitive performance on LIBERO-Spatial. The average score also presents substantial improvements, ranging from $4.7\%$ to $24.2\%$. These results further validate the effectiveness of \textbf{\textit{LiMoDE}} in multi-task learning. Although recent strong VLAs, such as $\pi_0$~\cite{black2410pi0}, CogACT~\cite{li2024cogact}, and OpenVLA-OFT~\cite{kim2025fine}, report higher multi-task performance, they differ substantially from us in architecture, model scale, pre-training data, and fine-tuning recipe. Tab.~\ref{liberomultitasks} is thus not intended to be a complete LIBERO leaderboard, but a compact auxiliary evaluation of the first pre-training stage.

\subsection{Ablation Studies}\label{abl}
\noindent\textbf{Key Components.}
In Tab.~\ref{ablation} (a), we evaluate the contribution of each key component and observe consistent performance improvements with the incremental integration of them. Applying MoE brings substantial gains, and incorporating the dynamics of experts further enhances the results. Based on it, the addition of heterogeneity in experts elevates MTFT and establishes stronger performance in OBJECT. Moreover, the sparse gating loss contributes significantly to spatial reasoning, while also advancing MTFT. Compared to learning task-specific LoRA, cross-task interaction by LiMoEAM demonstrates remarkable effectiveness, producing the most substantial single-component gain for MTFT and boosting all lifelong learning results, particularly in OBJECT. Finally, the complete model delivers the best results. Progressive improvements validate the complementary nature and the necessity of them.

\begin{table}[t]
\centering
\caption{\textbf{Study on key hyperparameters. Default configurations provide a stable and effective trade-off across the studied factors.}}\label{sensitivity}
\resizebox{0.9\linewidth}{!}{
\begin{tabular}{lcccccccccccc}
\toprule
\multirow{2}{*}{FWT ($\uparrow$)}
& \multicolumn{3}{c}{Task Number $N$}
& \multicolumn{3}{c}{Top-$k$ in LiMoEAM}
& \multicolumn{3}{c}{Rank in DyMoES}
& \multicolumn{3}{c}{Replay Epochs} \\
\cmidrule(lr){2-4} \cmidrule(lr){5-7} \cmidrule(lr){8-10} \cmidrule(lr){11-13}
& 2 & 6 & 10
& 1 & 2 & 3
& r$=$8 & r$=$16 & Hetero.
& 0\% & 10\% & 100\% \\
\midrule
OBJECT  & 84.0 & 82.0 & 81.0 & 77.0 & 81.0 & 79.0 & 78.0 & 79.0 & 81.0 & 77.0 & 79.0 & 81.0 \\
GOAL    & 73.0 & 71.0 & 70.0 & 66.0 & 70.0 & 68.0 & 67.0 & 68.0 & 70.0 & 67.0 & 68.0 & 70.0 \\
SPATIAL & 78.0 & 76.0 & 75.0 & 71.0 & 75.0 & 73.0 & 72.0 & 74.0 & 75.0 & 71.0 & 73.0 & 75.0 \\
\bottomrule
\vspace{-20pt}
\end{tabular}}
\end{table}

\begin{table}[!t]
\caption{\textbf{Computational efficiency comparison}. Our model achieves a trade-off between efficiency and performance.}
\label{cost}
\centering
\resizebox{0.90\linewidth}{!}{
\begin{tabular}{ccccccccc}
\toprule
\multirow{2}{*}{Model} & \multirow{2}{*}{Trainable/Total} & \multirow{2}{*}{FLOPS} & \multirow{2}{*}{Infer Time} & \multicolumn{3}{c}{Lifelong Learning (FWT)} \\
\cmidrule(lr){5-7}
& & & & OBJECT & GOAL & SPATIAL  \\
\midrule
TAIL~\cite{liu2023tail}         & 1.4/174.7M  & 297.4G & 121.9ms & 58.0 & 54.0 & 51.0 \\
IsCiL~\cite{lee2024incremental} & 12.7/187.3M & 423.3G & 135.7ms & 57.0 & 54.0 & 38.0 \\
LOTUS~\cite{wan2024lotus}       & 5.3/123.56M & 324.5G & 112.1ms & 74.0 & 61.0 & - \\
\cellcolor{pink!20}\textbf{Ours} & \cellcolor{pink!20}9.8/185.9M  & \cellcolor{pink!20}393.2G & \cellcolor{pink!20}177.8ms & \cellcolor{pink!20}\textbf{81.0} & \cellcolor{pink!20}\textbf{70.0} & \cellcolor{pink!20}\textbf{75.0}\\
\bottomrule
\vspace{-25pt}
\end{tabular}}
\end{table}

\noindent\textbf{Design Choices in DyMoES.}
The impact of design choices for DyMoES is studied in Tab.~\ref{ablation} (b). Replacing the token-wise MoE with a shared one results in performance degradation, indicating that the token-specific information is more important for different tasks. Instead, incorporating a task-independent $\bm{M}$ in expert selection still leads to a degradation in performance, indicating that considering motion in activation is beneficial. We further compare top-any activation with a Gumbel-Sigmoid alternative, where the latter performs worse, suggesting that our routing is more effective for dynamic expert selection. Moreover, a three-seed comparison with it shows comparable variance, indicating no noticeable training instability. Finally, to examine whether heterogeneous experts simply benefit from more parameters, we set the same rank $r_i$ to all experts, where the $r_i=16$ variant has a comparable total rank budget to our heterogeneous schedule. Results show that both homogeneous variants underperform our design, demonstrating that the improvement comes from capacity diversity rather than merely a larger parameter budget. These results validate the necessity of designs in DyMoES.

\begin{figure*}[t!]
\begin{center}
\includegraphics[width=0.9\linewidth]{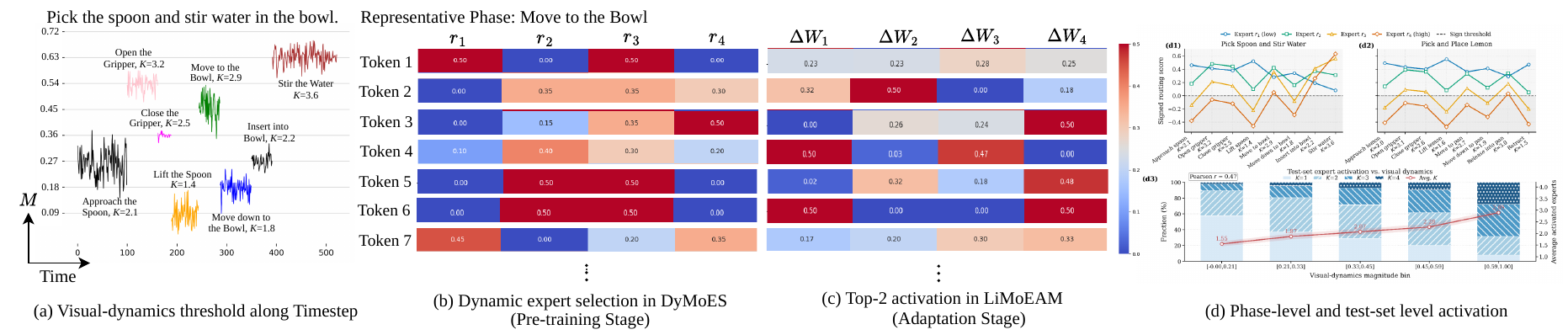}
\end{center}
\vspace{-10pt}
\caption{\textbf{Visualization of the visual-dynamics-conditioned threshold} and expert selection mechanisms. (\textbf{a}) The visual-dynamics-conditioned threshold $\bm{M}$ across timesteps for task ``Pick the spoon and stir water in the bowl''. The colored clusters denote the different manipulation phases. Each phase is also labeled with the average number of activated experts $K$, computed by averaging token-wise activations within that phase. (\textbf{b}) Token-wise dynamic expert selection in DyMoES for a representative phase (Move to the Bowl) during the multi-task pre-training stage. (\textbf{c}) Token-wise top-2 router activation in LiMoEAM during the lifelong adaptation stage. Darker red indicates a larger activation value. (\textbf{d1}) (\textbf{d2}) Phase-level routing activations of two tasks. (\textbf{d3}) Activated experts vs. motion magnitude across the real-world test set.}
\label{vis1}
\vspace{-5pt}
\end{figure*} 

\noindent\textbf{Design Choices in LiMoEAM.}
Tab.~\ref{ablation} (b) investigates design choices for LiMoEAM, where frozen Router means that the router is not updated during adaptation, and shared $\Delta W_n$ refers to adaptation with only one expert. All choices lead to negative impacts on the final results. Most notably, employing a frozen router during the adaptation fails to activate appropriate experts for cross-task interaction, leading to substantial performance deterioration. Moreover, excluding the current expert from computing the weight,~\textit{i.e.,} $C(x)_n$=1, prevents interaction between the current task and learned ones, resulting in suboptimal results. These observations highlight the role of selecting experts and cross-task interaction in adaptation.

\noindent\textbf{Sensitivity Analysis on Hyperparameters.}
Tab.~\ref{sensitivity} studies the sensitivity to different key hyperparameters. FWT gradually decreases as the sequence length $N$ increases, reflecting the greater difficulty of longer lifelong adaptation, while LiMoDE remains strong under the full 10-task setting. For LiMoEAM, top-2 activation achieves the best overall FWT by balancing cross-task interaction and avoiding irrelevant expert activation. For DyMoES, the heterogeneous Top-Any setting outperforms fixed-rank experts, showing the benefit of adaptive expert capacity. Increasing the coefficient replay epochs further improves FWT by stabilizing expert retrieval during lifelong adaptation. Overall, the default configuration provides a stable and effective trade-off across key hyperparameters.

\noindent\textbf{Computational Efficiency.}
Tab.~\ref{cost} summarizes the comparison. LiMoDE is built upon TAIL and introduces additional dynamic and lifelong experts for adaptive expert composition. Therefore, compared to it, LiMoDE naturally has more trainable parameters (8.4M) and moderately higher inference latency (55.9 ms). This moderate additional cost enables visual-dynamics-conditioned expert activation and lifelong expert composition, leading to substantial improvements in lifelong adaptation. Overall, LiMoDE requires 393.2G FLOPs and 9.8M trainable parameters, achieving a favorable performance-efficiency trade-off. Although LiMoDE is slower than LOTUS in raw inference latency, 177.8 ms vs. 112.1 ms, latency alone does not reflect the lifelong learning trade-off. In stead, our method substantially outperforms LOTUS and other baselines, especially on long-horizon manipulation tasks. Moreover, LiMoDE surpasses IsCiL in both SPATIAL performance and efficiency, demonstrating that strong lifelong robot learning can be achieved with moderate computational overhead.

\subsection{Visualization Analysis}\label{visu}
Fig.~\ref{vis1} visualizes the visual-dynamics-conditioned threshold $\bm{M}$, top-any activation in DyMoES, and the top-2 activation in LiMoEAM.
In Fig.~\ref{vis1} (a), the threshold $\bm{M}$ effectively captures the critical motion transitions during the stirring process, highlighting the temporal stages where primitive-level motion patterns dominate. Fig.~\ref{vis1} (b) shows how DyMoES dynamically selects experts based on motion cues, enabling the model to adaptively compose primitives from different skill domains. In contrast, 
Fig.~\ref{vis1} (c) shows that the top-2 routing in LiMoEAM activates the newly added expert ($\Delta W_4$) and one previously learned expert, combining them for the cross-task interaction. The visualization underscores the superiority of our motion-aware top-any gating and Lifelong task interaction in facilitating structured and interpretable skill reuse.

The phase-level routing is visualized in Fig.~\ref{vis1} (d1) and (d2). 
Each curve represents expert-wise averaged pre-sign routing scores across manipulation phases.
Clear patterns are observed. Smoother phases, such as lifting or transport, tend to activate fewer experts and are mainly associated with higher activation. In contrast, visually dynamic or contact-rich phases, such as opening the gripper and insertion, activate more experts, produce more positive routing scores, and stronger responses from higher-rank experts. 
These observations support our motivation that heterogeneous experts enable adaptive allocation for manipulation phases with different complexity.

Fig.~\ref{vis1} (d3) additionally verifies the behavior by visualizing $K$ \textit{vs.} visual-dynamics magnitude from the temporal difference of visual tokens over the real-world test set. It shows a clear positive trend (Pearson correlation $0.47$) that the average number of activated experts increases from $1.55$ to $3.08$. Low-dynamics regions are mainly associated with fewer activated experts, while higher-dynamics regions contain more activated experts. This is consistent with observations in Fig.~\ref{vis1} (d1) and (d2), further confirming the behavior of the dynamic router.


\section{Conclusion and Future Work}
This paper presented the \textbf{\textit{LiMoDE}} for lifelong robot manipulation. It had two key designs: a Dynamic Mixture of Expert Structure (DyMoES) to learn reusable skills and a Lifelong Mixture of Expert Adaptive Mechanism (LiMoEAM) to model task interaction, which efficiently transferred knowledge and mitigated catastrophic forgetting. Experimental results validated its effectiveness and efficiency. However, our framework still employs the top-k activation with experts of fixed rank in LiMoEAM. In the future, we will explore dynamic mechanisms to enable more flexible lifelong learning. While our current implementation uses a lightweight CLIP-based backbone, DyMoES and LiMoEAM operate on intermediate representations and can be integrated with stronger backbones, such as FlorenceVLM~\cite{chen2025florence}. We leave the systematic integration with such backbones to future work.
Finally, future work will explore more fine-grained visualizations of intermediate adaptation dynamics, such as adaptation efficiency.

{
    \small
    \bibliographystyle{IEEEtran}
    \bibliography{bare_jrnl_new_sample4}
}

\end{document}